\documentclass[letterpaper, 10 pt, conference]{ieeeconf}
\IEEEoverridecommandlockouts 
\usepackage {
  adjustbox,
  algorithm,
  algpseudocode,
  amsfonts,
  amsmath,
  array,
  booktabs,
  colortbl,
  enumitem,
  graphicx,
  lipsum,
  makecell,
  microtype,
  multirow,
  stfloats,
  subcaption,
  tabularx,
  textcomp,
  url,
  verbatim,
  xcolor,
  xspace,
}

\usepackage[T1]{fontenc}
\usepackage[hidelinks]{hyperref}
\usepackage{cleveref}

\AtBeginDocument{%
  }

\newcounter{myquestion}[section]

\newcounter{mytodo}[section]

\definecolor{mygreen}{RGB}{213,232,212}
\definecolor{myyellow}{RGB}{254,255,169}
\definecolor{myorange}{RGB}{254,212,169}
\definecolor{myred}{RGB}{255,97,97}
\definecolor{myblue}{RGB}{74,135,240}

\newcommand \thdr [1] {\textbf {#1}}

\newcommand \nfmd [1] {\texttt {#1}}

\floatname{algorithm}{Procedure}

\Crefname{section}{Section}{Sections}
\crefname{section}{Sect.}{sections}
\Crefname{table}{Table}{Tables}
\crefname{table}{Tbl.}{tables}
\Crefname{figure}{Figure}{Figures}
\crefname{figure}{Fig.}{figures}
\Crefname{equation}{Equation}{Equations}
\crefname{equation}{Eq.}{equations}

\title{\LARGE \bf NeRF-To-Real Tester: Neural Radiance Fields as Test Image Generators for Vision of Autonomous Systems}

\author{Laura Weihl$^{1}$, Bilal Wehbe$^{2}$ and  Andrzej W\k{a}sowski$^{3}$%
\thanks{$^{1}$Laura Weihl, IT University of Copenhagen, 2300 Copenhagen, Denmark 
{\tt\small lawe@itu.dk}}%
\thanks{$^{2}$Bilal Wehbe, Robotics Innovation Centre, DFKI, 28359 Bremen, Germany 
{\tt\small bilal.wehbe@dfki.de}}%
\thanks{$^{3}$Andrzej W\k{a}sowski, IT University of Copenhagen, 2300 Copenhagen, Denmark 
{\tt\small wasowski@itu.dk}}%
}

\begin{document}

\newcommand \nerfs {NeRFs\xspace}
\newcommand \nerf {NeRF\xspace}

\maketitle

\begin{abstract}
Autonomous inspection of infrastructure on land and in water is a quickly growing market, with applications including surveying constructions, monitoring plants, and tracking environmental changes in on- and off-shore wind energy farms. For Autonomous Underwater Vehicles and Unmanned Aerial Vehicles overfitting of controllers to simulation conditions fundamentally leads to  poor performance in the operation environment. There is a pressing need for more diverse and realistic test data that accurately represents the challenges faced by these systems. We address the challenge of generating perception test data for autonomous systems by leveraging Neural Radiance Fields to generate realistic and diverse test images, and integrating them into a metamorphic testing framework for vision components such as vSLAM and object detection. Our tool, N2R-Tester, allows training models of custom scenes and rendering test images from perturbed positions. An experimental evaluation of N2R-Tester on eight different vision components in AUVs and UAVs demonstrates the efficacy and versatility of the approach.
\looseness -1
\end{abstract}


\section{Introduction}%
\label{sec:introduction}

\emph{``I am not crazy; my reality is just different from yours.''} says the Cheshire Cat\,\cite{alice}. A roboticist might conclude that the Cheshire cat experiences the \emph{simulation-to-reality gap}; the gap between the confines of system's own objective reality, and the complexities of the real world. For Autonomous Underwater Vehicles (AUVs) and Unmanned Aerial Vehicles (UAVs), overfitting to simulation-based testing setups can lead to poor performance in the operating conditions of the real world. Testing research aims to decrease this gap.

Autonomous inspection of infrastructure, both underwater and on land, is a fast growing market. Applications include surveying constructions sites or wind farms for facilitating automatic detection of material degradation or environmental changes. To enable safe autonomous operations in real-world environments and sudden positional jumps due to adverse weather conditions such as wind or underwater currents, we need to establish the quality of different navigation systems and implementations, and ensure their reliability of establishing a consistent understanding of their environment.
\looseness -1

Integrating image-based perception into an autonomous navigation stack can enhance the AUV/UAV's real-time understanding of a scene for more fine-grained tasks. Light-weight Deep Neural Networks (DNNs) are often the default choice for tasks such as object detection or visual simultaneous localization and mapping (vSLAM). In aerial surveillance, accurate object classification and tracking are crucial for facilitating adaptive decision making. In underwater inspections, an Interest Point Detector (IPD) serves as the initial system to track the AUV's position in an open environment. Both image classifiers and IPDs are negatively affected by the adverse conditions such as bad visibility or sudden positional changes due to underwater currents or wind turbulence in UAV and AUVs respectively. 
\looseness -1

\begin{figure}[!t]
    \centering
    \includegraphics [width=0.8\linewidth, clip, trim = {5mm 2mm 10mm 1mm}]{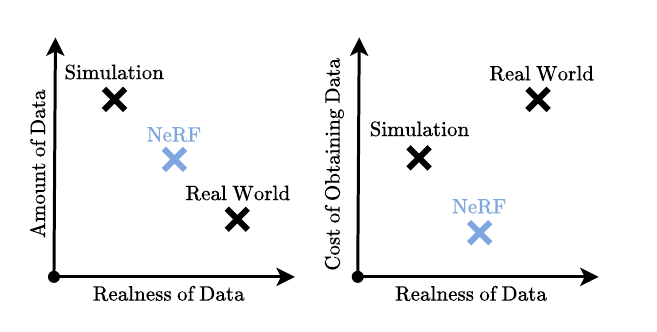}
    \caption{Trade-offs between the data realness and amount/cost.}%
    \label{drawio:info}
    \vspace* {-1 \baselineskip}
\end{figure}

For generating test data, simulators offer controlled environments and the ability to manipulate lighting conditions and scene characteristics. However, their effectiveness is constrained by the expertise and domain knowledge of the engineers who design them. A good performance on a simulation benchmark does not necessarily guarantee a good performance in the real world. Consequently, there is a pressing need for more diverse and realistic test data that accurately represents the challenges faced both by AUVs and UAVs. A trend in 3D scene reconstruction has emerged in recent years known as Neural Radiance Fields (NeRFs) \cite{mildenhall2021nerf}, providing a 3D representation of a scene in form of a neural network trained on camera poses and their respective images. NeRFs allow rendering novel views from the learned scene, by querying previously unseen camera poses. This positions NeRFs as a source of more realistic test data compared to engineered simulations, placing them at a sweet spot considering the trade-off when it comes to amount and cost of obtaining data (\cref{drawio:info}).
\looseness -1

We address the challenge of diversity and realness of perception test data for autonomous vehicles by leveraging NeRFs to generate realistic and diverse test images, with a specific focus on vision components such as vSLAM in AUVs and obstacle detection in UAVs. We adopt a metamorphic testing (MT) framework to uncover inconsistencies in several systems under test (SUT) using the \nerf-generated data. Specifically, we contribute:

\begin{itemize}[leftmargin=*]

  \item A short analysis of requirements and the existing methods for generating visual test data for autonomous systems (\cref{sec:requirements}),

  \item An adaptation of metamorphic testing for vision components in 3D space on real and NeRF-generated data (\cref{sec:method}),

  \item NeRF-to-Real-Tester (N2R-Tester), an implementation including NeRF training and image rendering\footnote{DOI: \href{https://zenodo.org/doi/10.5281/zenodo.14251863}{10.5281/zenodo.14251863}},

  \item An experimental evaluation of N2R-Tester on eight different vision components in AUVs and UAVs, demonstrating the efficacy and versatility of the approach (\cref{sec:evaluation,sec:results}).

\end{itemize}

To the best of our knowledge, we are the first to investigate how NeRFs can be applied in test image generation.
\section{Background}%
\label{sec:background}

\begin{figure}[t]
    \centering
    \includegraphics[width = 1.0\linewidth]{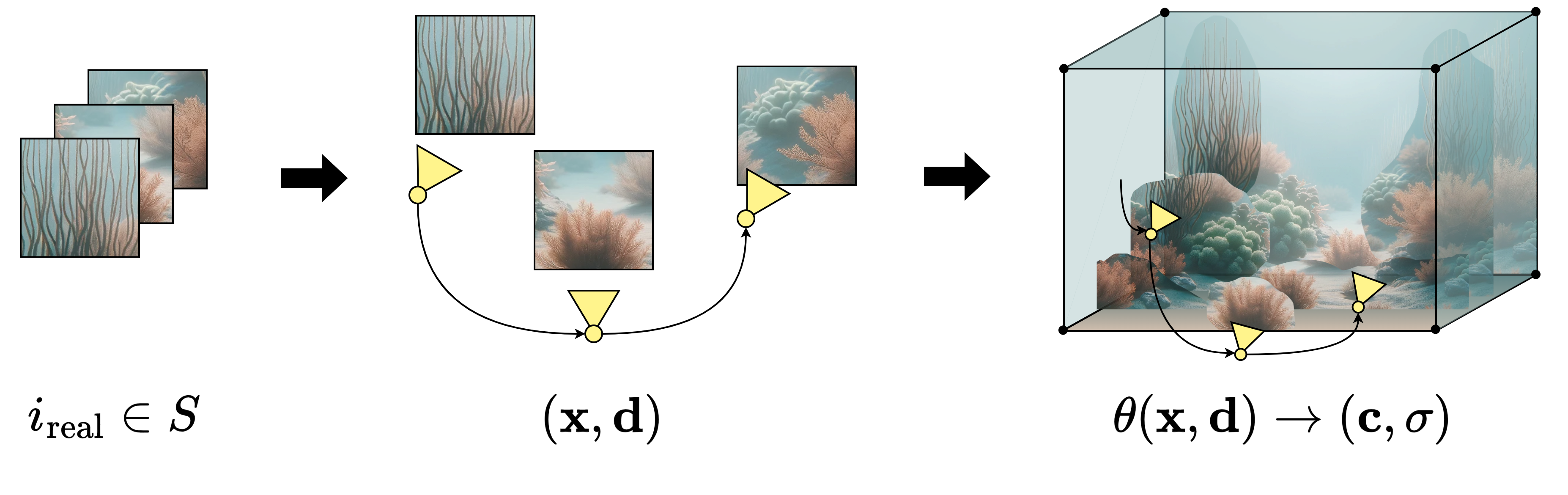}\\[-2mm]
    \textrm{\small NeRF Training}

    \includegraphics[width=1.0\linewidth]{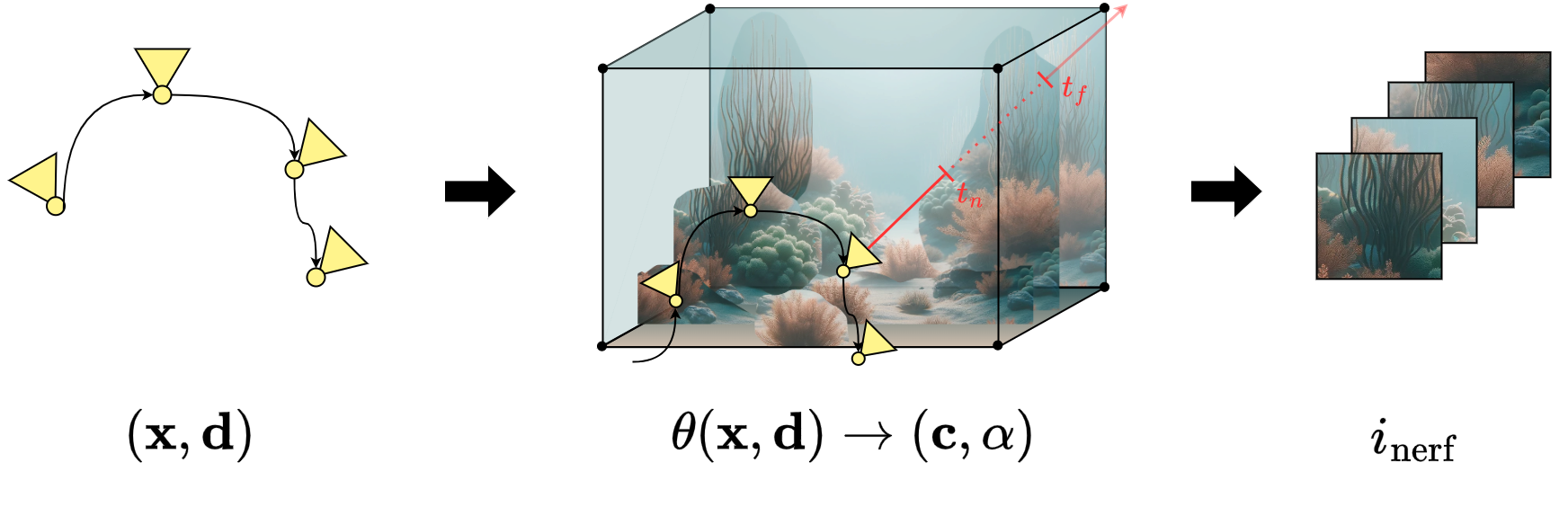}\\ 
    \textrm{\small NeRF Rendering}\\

    \caption{(Top) We take a set of real images of a scene \(i_\textrm{real} \!\in\! S\) as input. The NeRF model $\theta$ learns to map the estimated camera poses $\mathbf{x}, \mathbf{d}$ (yellow) to colour $\mathbf{c}$ and volumetric density  $\sigma$. (Bottom) We pass new camera poses $(\mathbf{x}, \mathbf{d})$ to the \nerf model $\theta$, creating  previously unseen views $i_\textrm{nerf}$ of the scene. Camera rays $\mathbf{r}(t)$ (red) are projected from the camera origin along the trajectory between a near and far bounds $t_n$, $t_f$.}%
    \label{fig:nerf-overview}
\end{figure}

Neural Radial Fields \cite{mildenhall2021nerf} (NeRFs) are function approximations of the form $\theta: (\mathbb{R}^3, \mathbb{R}^2) \rightarrow (\mathbb{R}^3, \mathbb{R})$. A NeRF learns a scene by mapping 3D camera locations $\mathbf{x}$ and viewing angles $\mathbf{d}$ to RGB color $\mathbf{c}$ and volume density $\sigma$, thus  $\theta (\mathbf{x}, \mathbf{d}) \rightarrow (\mathbf{c}, \sigma)$. A trained \nerf model is able to render novel views of a scene from previously unseen viewpoints, i.e. combinations of ($\mathbf{x}, \mathbf{d}$), as seen in \cref{fig:nerf-overview}. This is achieved by standard volumetric rendering techniques, i.e. a ray $\mathbf{r}$ is ``fired'' from the camera origin $\mathbf{o}$, as shown in \cref{fig:nerf-overview}. When tracing the ray between the near and far bounds $t_n$, $t_f$ and taking samples along the ray at different points $t$, the expected color $C(\mathbf{r})$ of this ray $\mathbf{r}(t) = \mathbf{o} + t\mathbf{d}$ starting at the camera origin $\mathbf{o}$ is given by:
\looseness -1

\begin{equation}
    C(\mathbf{r}) = \int^{t_f}_{t_n} \! \textrm{Tr}(t) \sigma (\mathbf{r}(t))\mathbf{c}(\mathbf{r}(t), \mathbf{d}) \, dt.
    \label{eq:sampling}
\end{equation}
The function $\textrm{Tr}(t)$ represents the probability of the ray traveling through $\sigma$ without being obstructed by any particle:
\begin{equation}
    \textrm{Tr}(t) = \exp \left(- \int^t_{t_n} \sigma (\mathbf{r}(s))ds \right).
\end{equation}

\section{What is a good test image?}%
\label{sec:requirements}

\looseness -1
For autonomous systems that operate in the real world, producing valid test images for perception is an active and challenging research area. It involves generating input images that preserve  photometric, semantic, geometric and other contextual information while introducing perceivable, realistic as well as diverse changes, for which we are still able to determine the expected test output. Especially for navigation applications such as vSLAM, spatio-temporal consistency is vital to preserve geometric information across consecutive frames under pose translations. We outline three key requirements for the generation of test images, that allow us to confidently assess the reliability of a system in real-world scenarios:

\begin{description}

  \item [Realness:] Generated test images are comparable to real scene images from the operational domain,

  \item [Diversity:] Generated test images add value to the available training data, so they differ from the training data,

  \item [Spatio-Temporal Consistency:] Consecutive images representing agent's motion present consistent geometry.

\end{description}

These are interconnected requirements on test image data that ensure a robust and comprehensive testing of perception systems for navigation. Without realness we cannot confidently assess the reliability of a system in real-world scenarios. Without diversity we cannot test generalization. Without spatio-temporal consistency, testing of a navigational systems is void.

\paragraph{Realness}
Strictly speaking, valid test inputs for testing perception are the images of real world scenes captured by a camera. However, using only real world scene data is often too costly. For instance, in underwater robotics, acquiring data necessitates offshore missions, involving considerable resources and time. Therefore, alternative methods are necessary to obtain additional yet realistic images. One possibility is to resort to simulation\,\cite{alvarez2023mimir}. The images produced by simulators often possess an artificial visual quality and intensify the simulation-to-reality gap. Alternatively, adversarial methods create imperceptibly small noise vectors that, when added to a real image, push it over a decision boundary\,\cite{goodfellow2014explaining}. Similarly, coverage-based methods that generate images triggering faulty behavior, are inherently adversarial, directly attacking a model based on its internal learned representations \cite{pei2017deepxplore, tian2018deeptest}. We treat the perception systems as black-box and therefore provide a more generalizable method.
\looseness -1

\paragraph{Diversity}
Test data generation aims to extend beyond the training data, while maintaining realness. Simple methods like rotating and scaling images, known as data augmentation, are common in DNN training. Most of the existing research on the diversity of the generated data focuses on more advanced transformation techniques and on measuring realness or validity of the generated inputs. One group of methods patches existing images; pasting new objects and changing properties of objects (e.g.\ lighting or color\,\cite{woodlief2022semantic}). However, these can be easily identified as unreal by humans. A less direct approach is to perform search in the feature space\,\cite{zohdinasab2021deephyperion}, or to change global properties of an image, like weather conditions, by use of Generative Adversarial Networks (GANs) \cite{zhang2018deeproad}. A GAN incorporates a discriminator network which optimizes the realness of the generated images. Crucially, the discriminator network learns statistical patterns and not the semantic content in the generated images. It might learn to exploit weaknesses in the discriminator without actually learning the representation of the underlying distribution. Unlike the above, we use transformations in the \nerf space (rolls and translations) to create new inputs for the perception components. Instead of challenging the SUT with new object and lighting configurations, we generate new view angles and new occlusions.
\looseness -1

\paragraph{Spatio-temporal Consistency}

For testing perception in the context of navigation, one needs to produce image sequences that differ in the camera location, and that manifest the corresponding changes to the geometry of objects in the scene-consistent variation of positions, view angles, and occlusions. \nerfs were originally developed precisely for this requirement. Unlike GANs, where the learned distribution is a complex manifold in a high-dimensional space, giving poor control in the image generation process, \nerfs can render new images parameterized with camera positions in Euclidian space. This allows to emulate the motion of the agent, while maintaining the realness driven by the training data in the scene.
\looseness -1
\section{Method}%
\label{sec:method}

We describe our method to  use NeRFs as test image generators within a customized metamorphic testing procedure.

\begin{figure}[t]

  \begin{center}
    \includegraphics [
      width = 0.75 \linewidth,
      clip,
      trim = 0mm 0mm 0mm 0mm
    ]{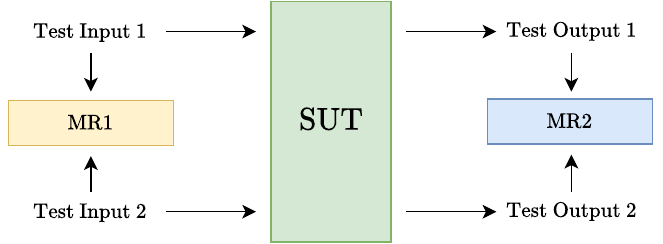}
  \end{center}

  \caption{Metamorphic testing: Metamorphic relations (MRs) are established between pairs of test inputs and between the corresponding outputs. The test fails if MR2 does not hold.}%
  \label{figure:mt-basic}

\end{figure}

\subsection{Metamorphic Testing}

\begin{figure}[t]
  \begin{center}
    \includegraphics [
      width = \linewidth, clip, trim = 0mm 2.5mm 0mm 2.5mm,
    ]{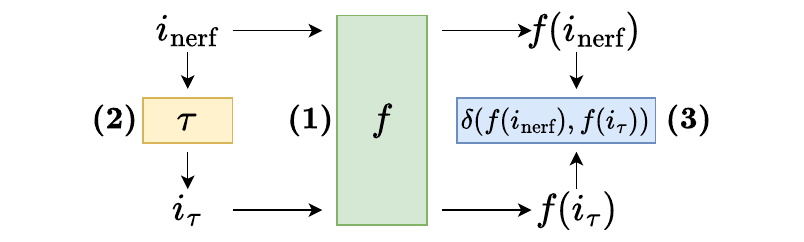}
  \end{center}

  \vspace {-1mm}

  \caption{N2R-Tester: \textbf{(1)} an image processing SUT $f$, \textbf{(2)} a metamorphic relation (MR) in form of a pose transformation $\tau$ rendered in a NeRF model, \textbf{(3)} an MR $\delta$ checking inconsistencies between the SUT outputs $f(i_\textrm{nerf})$ and $f(i_\tau)$.}%
  \label{figure:mt-nerf}
\end{figure}

Metamorphic Testing (MT) is a testing framework to assess the accuracy and robustness of a System Under Test (SUT)\,\cite{chen2018metamorphic}. It is particularly well-suited when the quality of the expected output of a SUT is difficult to establish due to high dimensionality of the test data or absence of ground truth. Unlike traditional testing methods that rely on input-output pairs, in MT we formulate metamorphic relations (MRs) between input-input and output-output pairs that the SUT is expected to comply with (\cref{figure:mt-basic}). MT therefore aims to capture the intrinsic properties of the data that the system processes. While other testing techniques aim at detecting bugs on a code level, MT can uncover critical inconsistencies on a behavioral level of a SUT by observing discrepancies in test outputs, given interrelated test inputs. MT is highly suitable for testing image-based perception systems where the inherent complexity and unstructured nature of image data poses significant challenges.\@ Given a SUT $f$ that processes images $i$ of a scene $S$, a metamorphic relation can be expressed as $ \forall i \in S. \;  f(g_I(i)) = g_O(f(i))$. Here, \(g_I, g_O\) are functions that transform the program input and output domain of $f$ respectively. We transform the test images $i$ according to function $g_I$ to generate new tests and validate them against the output of the SUT $f(i)$ by function $g_O$. In this work, we utilize NeRFs as automatic test image generators $g_I$ while simultaneously monitoring the effects that NeRF-generated images have on the output of a SUT (\cref{figure:mt-nerf}). To achieve this we employ a range of transformations of an input image as our input MR and evaluate discrepancies between the outputs of the SUT for the transformed images using critical performance metrics. The strength of \nerfs lies in their ability to render realistic images from previously unseen camera poses \((\mathbf{x}, \mathbf{d})\) of a scene or object, while preserving its photometric, structural and geometric information, which simultaneously offers image diversity by giving different viewpoints. We exploit this ability by applying a range of transformations $T$ to $(\mathbf{x}, \mathbf{d})$ in the 3D NeRF space. We denote the a transformed image $i_{\tau}$. We report how a SUT $f$ responds to a range of diverse test images by the changes in MR outputs, i.e.
\( \delta ( f(i_\textrm{nerf}), f(i_\tau) ) \). The predicate INC captures the violation written using a performance metric \(q\) into real numbers: 
\looseness -1

\begin{equation}
    \textrm{INC}(f) = q (f(i_\textrm{nerf}), f(i_{\tau}))  > \epsilon.
\end{equation}
\looseness -1

\noindent

\subsection{NeRF-To-Real Tester}

N2R-Tester \(\langle f, S, T, \delta \rangle\) comprises a system under test $f$, a set of images $S$ of a scene or object, a list of pose transformations \(T\), and metamorphic relations $\delta$. N2R-Tester takes the image set $S$ and reconstructs a 3D scene solely based on these 2D input images in form of a NeRF model. We achieve this by feeding all images $i \in S$ to a structure-from-motion tool, colmap\,\cite{schonberger2016structure}, which selects a subset of all images $S_\textrm{real} \subseteq S$ to get estimated camera poses $(\mathbf{x}, \mathbf{d})$ for each image in $i_\textrm{real} \in S_\textrm{real}$. These image/camera-pose combinations are used to train a NeRF model $\theta$. Now we render the same camera poses in the NeRF model to get all corresponding NeRF-generated images $i_\textrm{nerf}$ from the identical viewpoints. Subsequently, we apply a number of different transformations $\tau \in T$ to each of the camera poses $(\mathbf{x}, \mathbf{d})$ to render the transformed images in the NeRF model to get $i_\tau$. The supported transformations are translations of position as well as rotations in roll, pitch and yaw angle. 
\looseness -1

\section{Experimental Evaluation}%
\label{sec:evaluation}

We evaluate N2R-Tester experimentally, in two application scenarios: (1) testing an interest point detector of an AUV during a wall inspection mission, and (2) testing an image classifier in an UAV during a vehicle inspection mission. Both scenarios require operating in a 3D space with 6 degrees of freedom, i.e. in air and in water, as well as their exposition to air or water turbulence causing abrupt positional jumps which challenge the understanding of the scene. Specifically, we pose the following research questions (RQs).

\begin{itemize}[leftmargin=*]

    \item{\textbf{RQ1}:} How effective is N2R-Tester at generating \emph{realistic} images of a scene or object as perceived by an image processing SUT?

    \item{\textbf{RQ2}:} How effective is N2R-Tester at generating \emph{diverse} images of a scene or object that uncover inconsistent behaviors in an image processing SUT?

    \item{\textbf{RQ3}:} How \emph{efficient} is N2R-Tester at detecting inconsistencies in the behavior of an image processing SUT?

\end{itemize}

Research questions RQ1 and RQ2 address the two core requirements of \cref{sec:requirements}.  The spatiotemporal consistency is satisfied by design, and it is evaluated in \nerf papers.

\subsection{Training Data}

\paragraph {AUV Wall Inspection}

We record five videos of underwater scenes using a manually operated underwater drone equipped with a single front-facing camera. We mimic a trajectory of an AUV during inspections of vertical planar walls. Each video lasts approximately one minute. Images are extracted at 25 FPS and the resolution of \(1920 \times 1080\).  The scenes contain walls covered by various types of marine growth. We estimate camera poses with colmap\,\cite{schonberger2016structure} (\cref{tab:data_info}).
\looseness -1

\noindent \paragraph {UAV Vehicle Inspection} We use two image data sets available in nerfstudio \cite{tancik2023nerfstudio} (\nfmd{plane}, \nfmd{dozer}), captured with a handheld camera device. We record three additional scenes with a handheld camera, showing a car, a truck and a bike and extract images at 25 FPS with image resolution of $1920 \times 1080$ pixels.  We use colmap for camera pose estimation. The images of all data sets have the respective vehicle roughly at the image center and the camera trajectory mimics a UAV circling the vehicle at an approximately fixed level above ground. See also \cref{tab:data_info} for details on the trained NeRF models.
\looseness -1

\begin{table}[t]

    \renewcommand \arraystretch {1.0}

    \begin{tabularx}{\linewidth}{
      >{\small}X
      >{\small}c
      >{\small\raggedright}p{15mm}
      >{\small\raggedleft}p{10mm}
      >{\small\raggedleft}p{10mm}
      >{\small\raggedleft\arraybackslash}p{12mm}
    }
        \thdr{\nerf Model}
        & \thdr{Mission}
        & \thdr{Image Resolution}
        & \thdr{\#Total Images}
        & \thdr{\#Train Images}
        & \thdr{Train Time(h)}
        \\\hline  \\[-2mm]

        \nfmd{dory1}   & AUV   & 1920$\times$1080   & 301  & 271  & 1.38 \\
        \nfmd{dory2}   & AUV   & 1920$\times$1080   & 330  & 297  & 1.88 \\
        \nfmd{dory3}   & AUV   & 1920$\times$1080   & 358  & 323  & 1.39 \\
        \nfmd{dory4}   & AUV   & 1920$\times$1080   & 303  & 273  & 1.52 \\
        \nfmd{dory5}   & AUV   & 1920$\times$1080   & 356  & 321  & 1.41 \\
        \nfmd{plane}   & UAV   & 1080$\times$1920     & 317  & 286  & 1.68 \\
        \nfmd{dozer}   & UAV   & 3008$\times$2000     & 359  & 324  & 1.38 \\
        \nfmd{car}     & UAV   & 1920$\times$1080     & 532  & 479  & 1.28 \\
        \nfmd{truck}   & UAV   & 1920$\times$1080     & 418  & 377  & 1.39 \\
        \nfmd{bike}    & UAV   & 1920$\times$1080     & 303  & 273  & 1.31 \\

    \end{tabularx}


    \caption{Details of NeRF models: All models are our own except \nfmd{plane}, \nfmd{dozer}, available in nerfstudio.}
    \label{tab:data_info}

\end{table}

\subsection{Systems Under Test}

\paragraph {Interest Point Detectors (AUV)}

We evaluate N2R-Tester on four AUV components as SUTs: two DNN-based IPDs as well as two popular state-of-the-art non-DNN IPDs. We use the OpenCV implementations of SIFT and ORB \cite{bradski2000opencv}. SuperPoint with pre-trained weights is available on github \cite{detone2023superpointgithub}. UnSuperPoint is proprietary software and currently not publicly available with pre-trained weights.
\looseness -1

\paragraph {Image Classifiers (UAV)}

We evaluate N2R-Tester on four UAV components as SUTs; two small DNNs with $\sim$4-5M parameters, selected for their capacity to be deployed on modern quadrocopters, as well as two large DNNs. We choose MobileNet\,, EfficientNet, XCeption, and VGG16, all as implemented in Keras\,\cite{chollet2015keras}. All models have a convolutional architecture and are trained on ImageNet-1k\,\cite{deng2009imagenet}.
\looseness -1

\subsection{Image Quality Metrics}
We consider the following image quality metrics, when comparing an image to a reference image:
\begin{itemize}[leftmargin=*]
    \item{Peak-Signal-To-Noise Ratio (PSNR)}: based on mean squared error, scaled to pixel ranges of the underlying distribution.
    \item{Structural Similarity Index (SSIM)}: estimates similarity according to luminance and contrast \cite{wang2009mean}. 
    \item{Learned Perceptual Image Patch Similarity (LPIPS)}: measures similarity in a DNN feature representation \cite{zhang2018unreasonable}.
\end{itemize}

\subsection{Performance Metrics for Systems Under Test}

We consider five SUT metrics, three for the UAV mission and two for the AUV mission, and inject them into N2R. 

\begin{itemize}[leftmargin=*]
    \item{Image Classification: Cosine Similarity $\uparrow$}: cosine angle between two DNN outputs; two parallel vectors have a cosine similarity of $1$, two orthogonal vectors of $-1$. DNN outputs are the probability distributions over 1000 classes.
    
    \item{Image Classification: L2 Norm $\downarrow$}: Euclidean squared distance between the two DNN output vectors with minimum value of 0 and the maximum value of $\sqrt{2}$.

    \item{Image Classification: Class Invariance $\uparrow$}: equivalence of the class with the highest probability of two DNN outputs. 
    
    \item{IPD: Repeatability Score $\uparrow$}: ratio of matched points by projecting one set of points into a second image with a homography \cite{schmid2000evaluation}, with maximum discrepancy of 2 pixels.

    \item{IPD: Interest Point Spread $\uparrow$}: ratio between the intersection and union of the coverage areas of interest points, demonstrating point stability \cite{bailo2018efficient}. 
    
\end{itemize}

\subsection{NeRF Models}

The structure-from-motion tool colmap selects a minimum 300 suitable images of a scene by balancing sufficient visual information and processing speed. Colmap subsequently estimates the camera poses for each image. For our NeRF models we choose a network architecture called Nerfacto, available in nerfstudio \cite{tancik2023nerfstudio} because of its balance between  training/rendering speed and visual quality. We train a total of ten \nerf models on underwater and aerial scenes listed in \cref{tab:data_info}. We mainly use the default parameters in nerfstudio with 30k training steps and 4096 ray samples per step.
\looseness -1

\subsection{Evaluation for N2R-Tester}

To demonstrate the efficacy of N2R-Tester in producing realistic and diverse test images, and thus answering RQ1, RQ2, and RQ3, we inject our SUTs, NeRF models and MRs into our framework. The MRs between test inputs are the chosen spatial transformations $\tau$. The MRs between test outputs are the changes on SUT performance on metric $q$, that is $q ( f(i_\textrm{nerf}), f(i_\tau) ) > \epsilon $, with three threshold values for $\epsilon$. Additionally, we employ artificial image mutations suggested by authors of DeepXplore \cite{pei2017deepxplore} to enable a baseline comparison. Note that in their work, the image mutations are used to generate test images to find DNN test inputs that trigger differential SUT behavior and cause high neuron coverage. We are not interested the neuron coverage metric since it has shown inconclusive efficacy in other works \cite{yang2022revisiting}. We still choose the DeepXplore mutations because they are engineered artificially, thus we are curious to their level of realism and comparative output in SUT behavior. Since each NeRF model has its own scale, we design targeted pose transformations for each NeRF model individually. We consider the reality-to-NeRF domain shift a transformation in itself and call it $\tau_0$. We perform six types of pose transformations in the NeRF model:

\begin{itemize}[leftmargin=*]
    \item{$\tau_0$}: reality-to-NeRF domain shift,
    \item{$\tau_1$}: small translation on x-axis, small rotation in yaw,
    \item{$\tau_2$}: small translation on y-axis, small rotation in pitch,
    \item{$\tau_3$}: large translation on x-axis, large rotation in yaw,
    \item{$\tau_4$}: large translation on y-axis, large rotation in pitch,
    \item{$\tau_5$}: transformation (1), small roll rotation,
    \item{$\tau_6$}: transformation (2), small roll rotation.
\end{itemize}

\begin{figure}[ht]
\centering
  \begin{subfigure}{0.32\linewidth}
    \includegraphics[width=\linewidth]{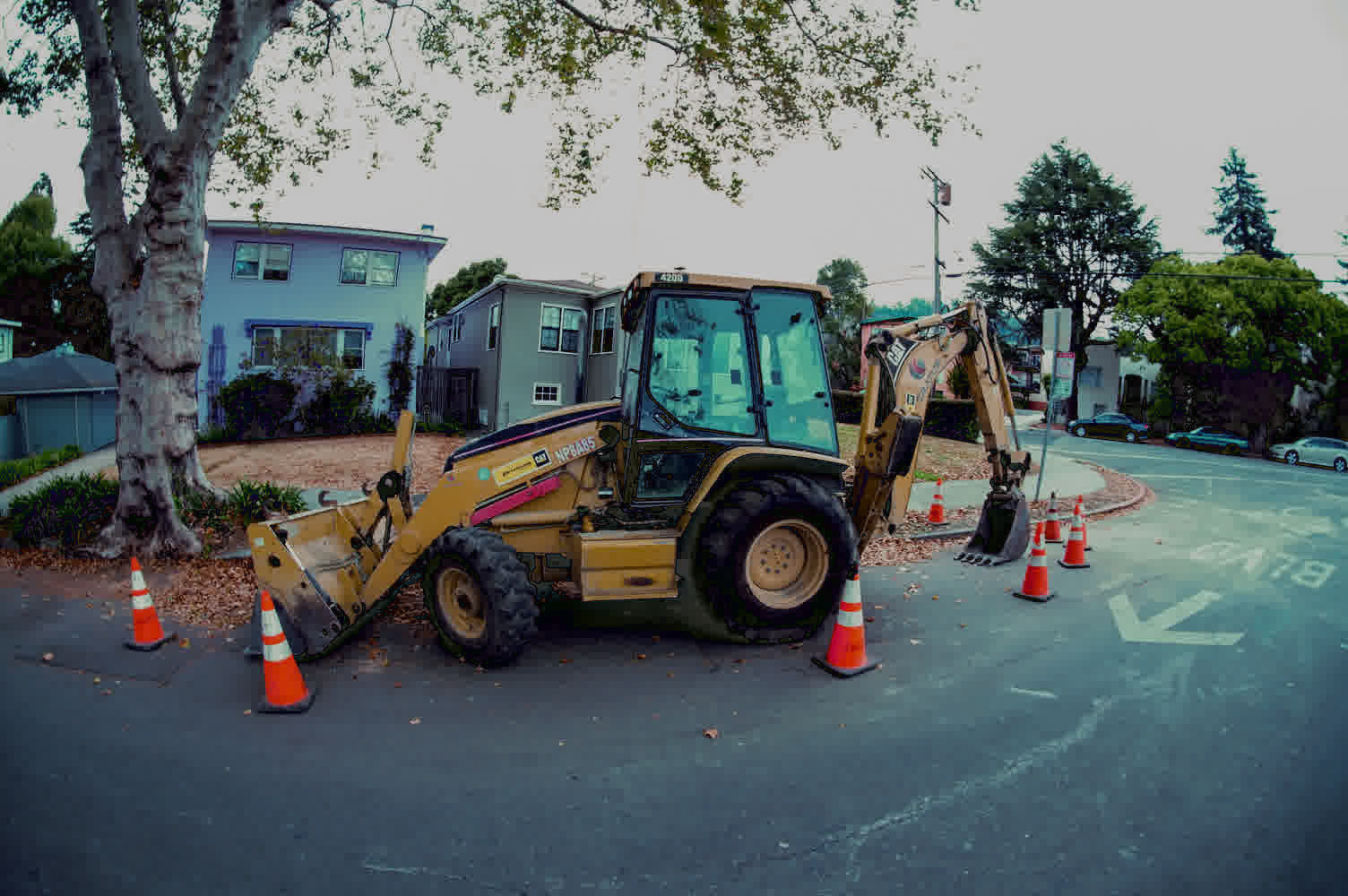}
     \caption{$m_1$}
  \end{subfigure}
  \begin{subfigure}{0.32\linewidth}
    \includegraphics[width=\linewidth]{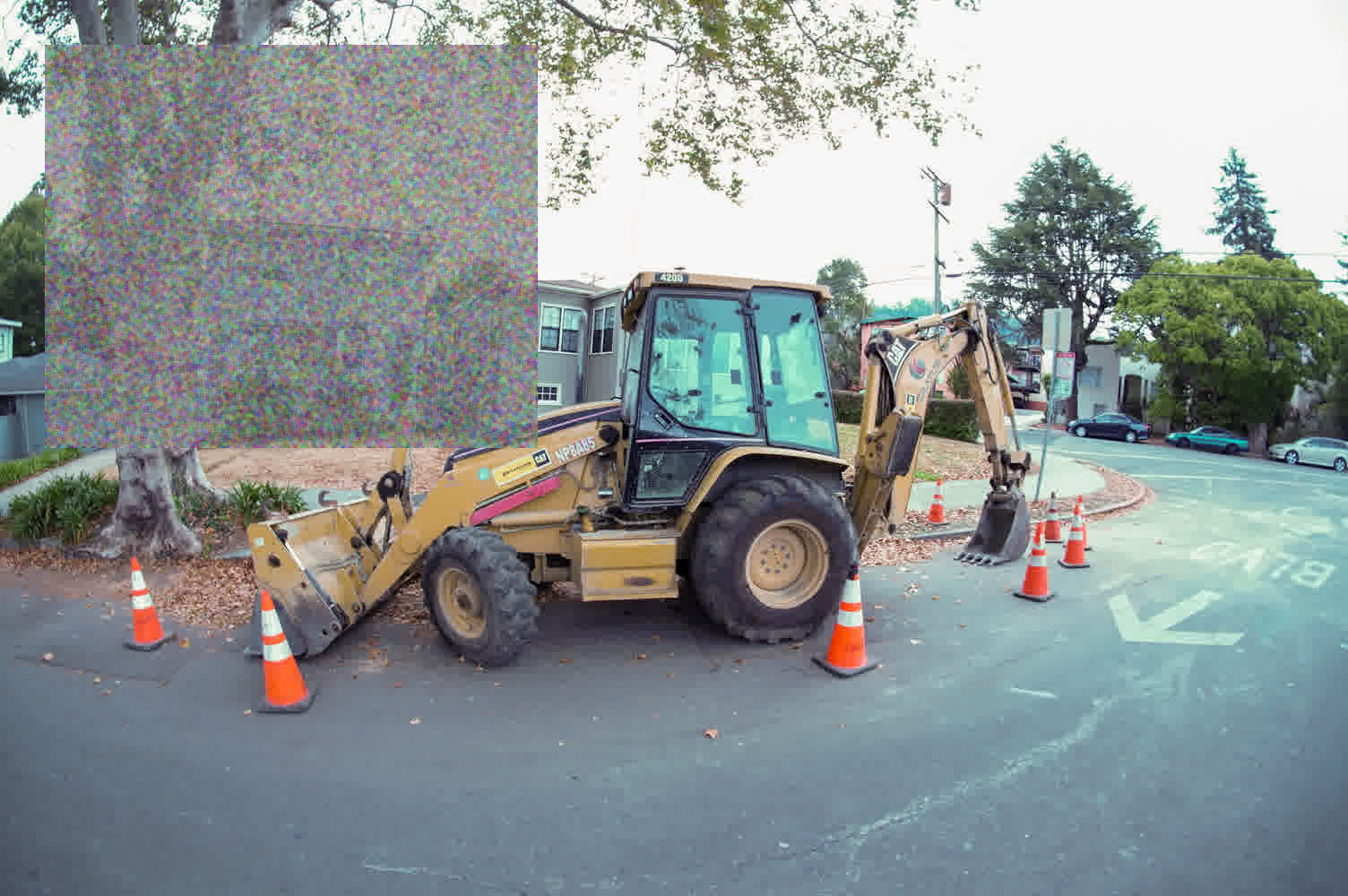}
     \caption{$m_2$}
  \end{subfigure}
  \begin{subfigure}{0.32\linewidth}
    \includegraphics[width=\linewidth]{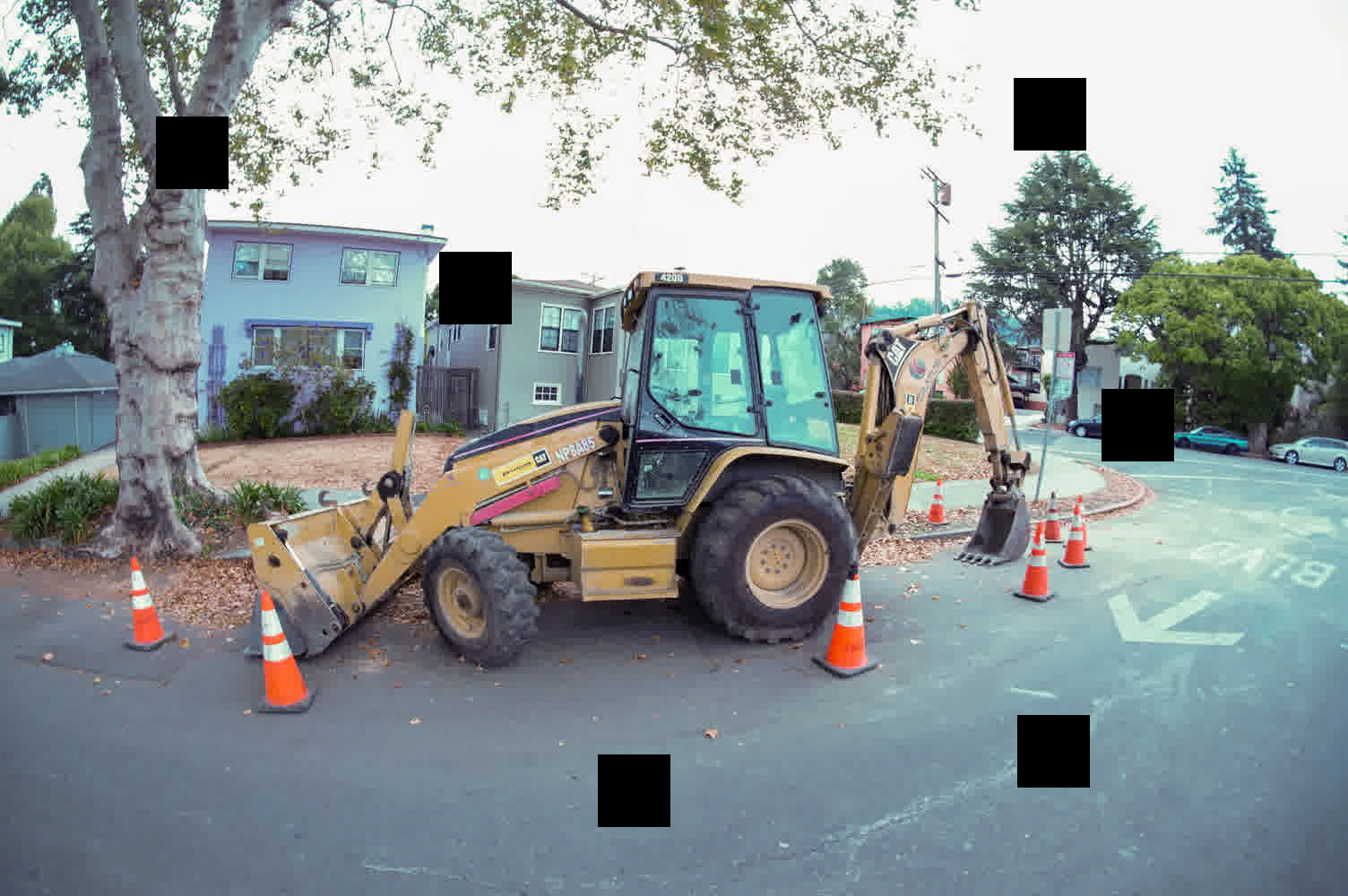}
     \caption{$m_3$}
  \end{subfigure}
    \caption{DeepXplore mutations of real images of a dozer, with (a) changes in brightness, (b) a random pixel patch and (c) multiple black patches, imitating camera failure modes.}
  \label{fig:rq2-deepxplore}

\end{figure}

\noindent
We manually design the transformations, such that the main object or viewpoint of the scene is in full view at all times. Therefore the image classifier should return the same object class at all camera poses. Similarly the IPD should give consistent interest points under our transformations. To achieve this, for example when translating the camera upward, we apply a small rotation in pitch downward. By applying these transformations, we multiply the amount of test images in each scene sixfold. Note that the pairs of transformations $\tau_1$/$\tau_2$, $\tau_3$/$\tau_4$) and $\tau_5$/$\tau_6$ increasingly deviate from the original camera path and should therefore produce more diverse images. We choose three image mutations by DeepXplore as our baseline, because from a practical point of view, they come closest to real-life failure modes in cameras (examples in \cref{fig:rq2-deepxplore}):
\begin{itemize}[leftmargin=*]
    \item{$m_1$}: changes in image brightness, e.g. caused by automated camera over-/underexposure,
    \item{$m_2$}: applying one large patch of noisy pixels, e.g. caused by camera malfunction, transmittance failures or loose cables,
    \item{$m_3$}: applying six small patches of black pixels, e.g. caused by dirt or stains on the camera lens.
\end{itemize}

\section{Results}%
\label{sec:results}

\begin{figure}

  \begin{subfigure}{0.48\linewidth}
    \includegraphics[width=\linewidth]{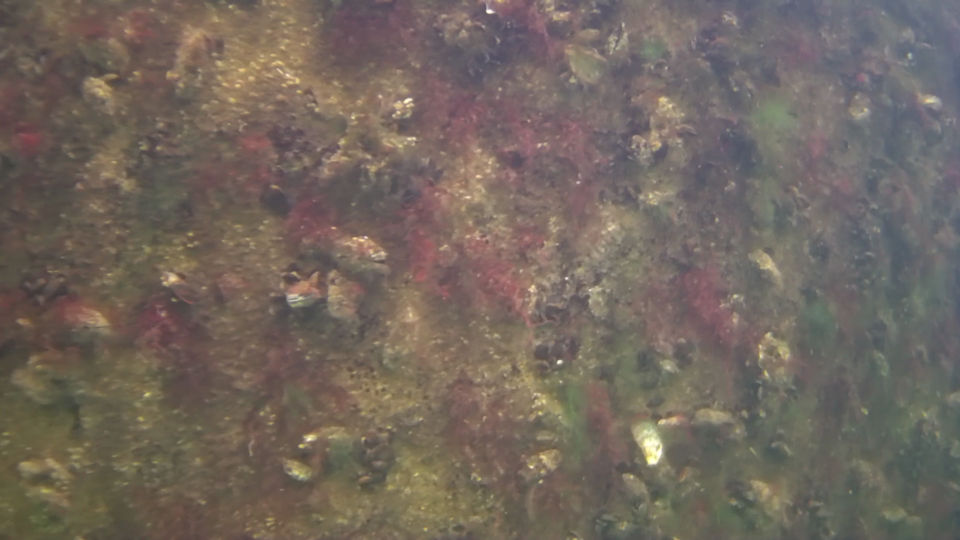}
    \caption{Real image}
  \end{subfigure}
  \hfill 
  \begin{subfigure}{0.48\linewidth}
    \includegraphics[width=\linewidth]{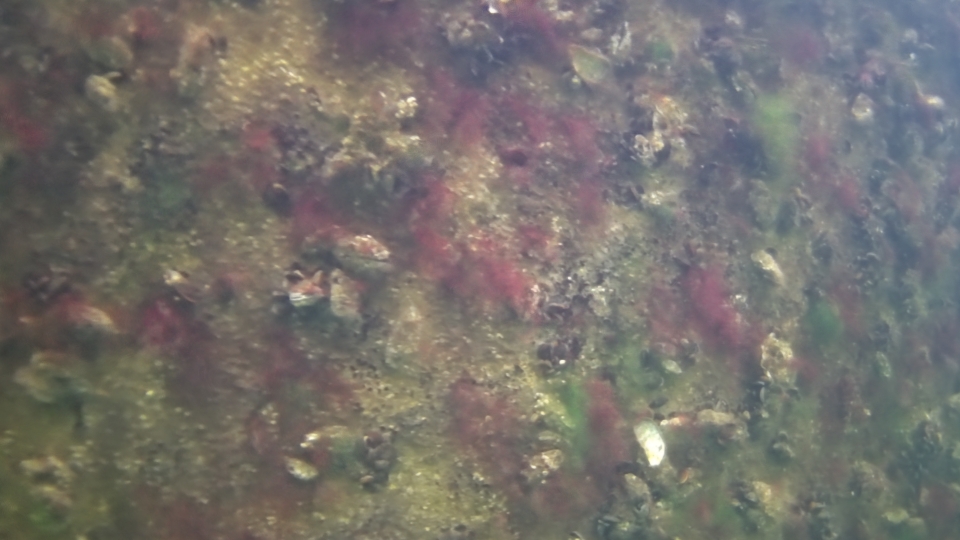}  
    \caption{N2R $\tau_0$}
  \end{subfigure}
  \begin{subfigure}{0.48\linewidth}
    \includegraphics[width=\linewidth]{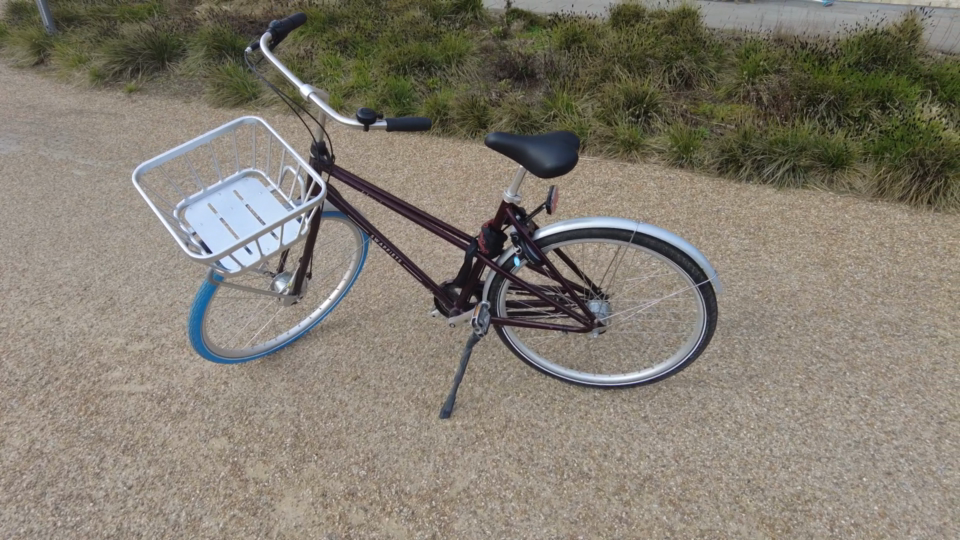}
    \caption{Real image}
  \end{subfigure}
  \hfill 
  \begin{subfigure}{0.48\linewidth}
    \includegraphics[width=\linewidth]{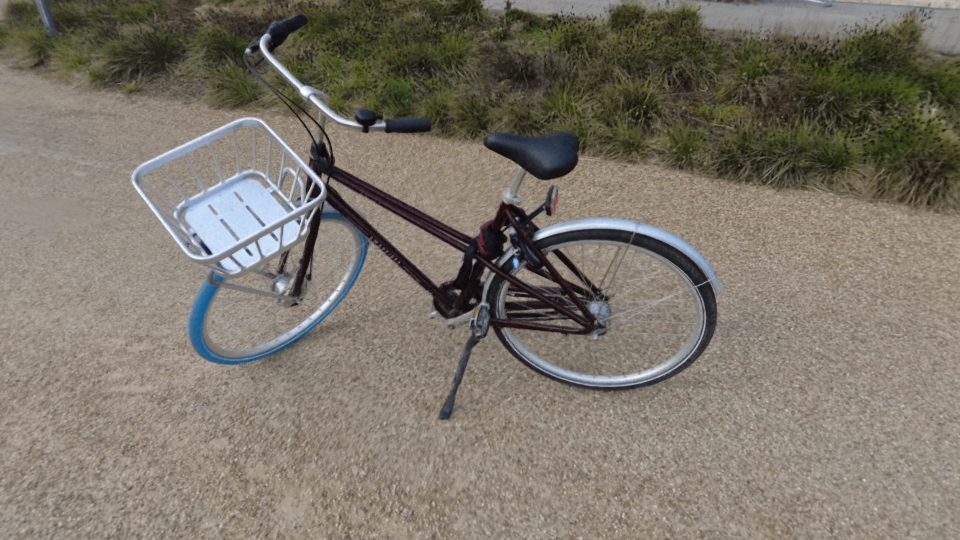}
     \caption{N2R $\tau_0$}
  \end{subfigure}

  \caption{(a) A real image of an underwater scene of an AUV mission and (b) its pose-equivalent image in NeRF model \nfmd{dory1}. (c) A real image of an UAV mission, and (d) its pose-equivalent image in NeRF model \nfmd{bike}.}
  \label{fig:real-vs-nerf}
    
\end{figure}

\begin{figure*}[htbp]
  \centering
  \begin{subfigure}{0.25\textwidth}
    \includegraphics[width=\linewidth]{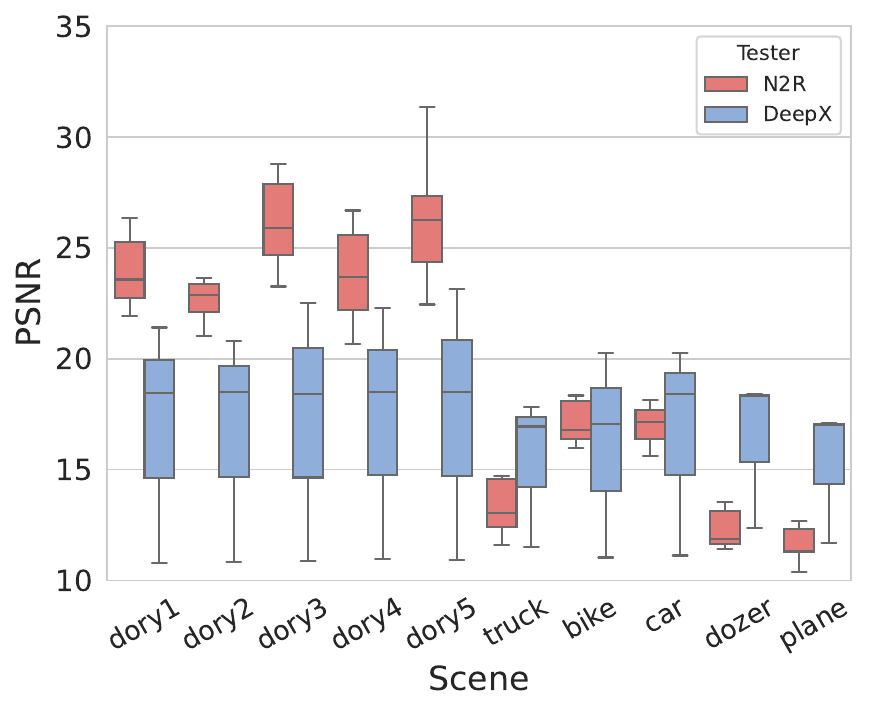}
    \caption{Avg. PSNR $\uparrow$}
  \end{subfigure}
  \begin{subfigure}{0.25\textwidth}
    \includegraphics[width=\linewidth]{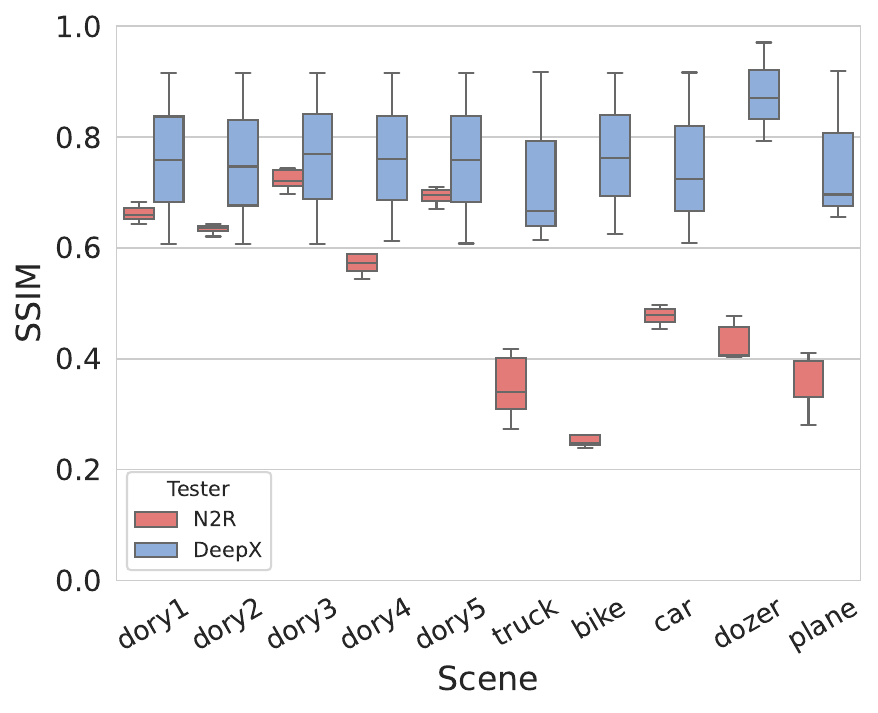}
    \caption{Avg. SSIM $\uparrow$}
  \end{subfigure}
    \begin{subfigure}{0.25\textwidth}
        \includegraphics[width=\linewidth]{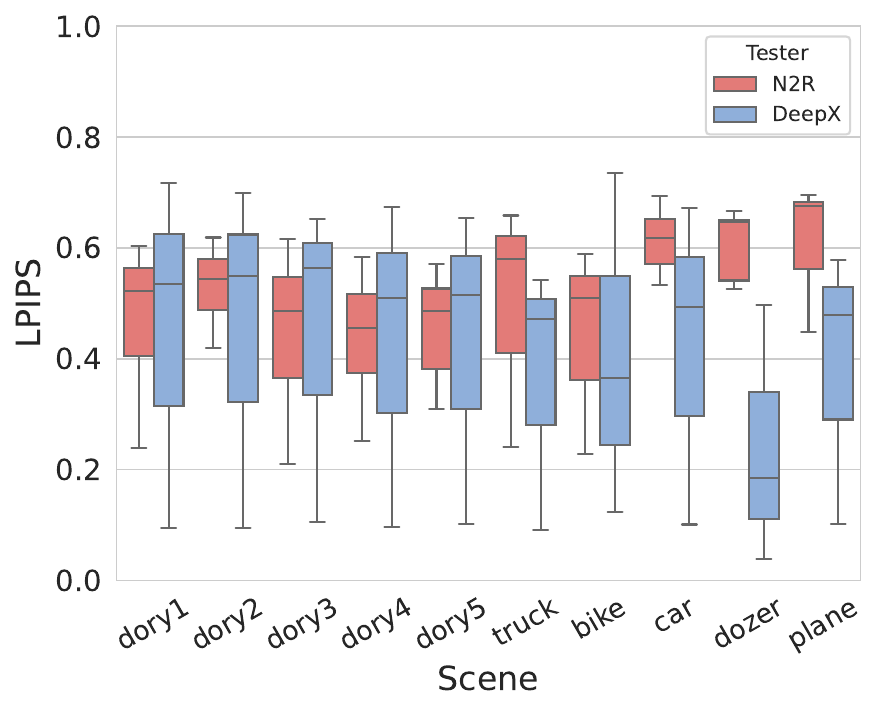}
    \caption{Avg. LPIPS $\downarrow$}
    \end{subfigure}
  \caption{Image metric values for PSNR, SSIM, and LPIPS of images generated by N2R (red) vs DeepXplore (blue).}%
  \label{fig:rq1-all-image-metrics}
\end{figure*}

\noindent
In this section, we discuss the experiment results of N2R-Tester, and its ability to test our SUTs, i.e. IPDs and image classifiers. All test images are rendered at a resolution of 960x540 pixels.

\subsection{RQ1: Realness}

\paragraph {Visual quality of NeRF-generated images}

\Cref{fig:real-vs-nerf} shows real and NeRF-generated images of underwater and vehicle scenes to convince the reader of the visual quality of NeRF-generated images. The image generated by the underwater model \nfmd{dory1} demonstrate an almost indistinguishable resemblance with the real world scene. The image produced by \nfmd{bicycle} shows a similar visual quality and convince by subtle details, e.g. the bicycle spokes. We found the near and far bound parameters $t_n$, $t_f$ in \cref{eq:sampling} supported convergence during the reconstruction process for \nfmd{dory1} and \nfmd{dory3} based on evaluation loss. We report the image quality metrics PSNR, SSIM and LPIPS for each scene in \cref{fig:rq1-all-image-metrics}. N2R images of the underwater models show better values for PSNR and LPIPS than DeepXplore mutations, suggesting a higher level of realness. Under closer investigation, $\tau_0$ and $m_3$ show the least amount of visual change, and also show the best image quality metrics, as expected.
\looseness -1

\begin{figure*}[hbt!]
  \centering
    \begin{subfigure}{0.19\textwidth}
    \includegraphics[width=\linewidth]{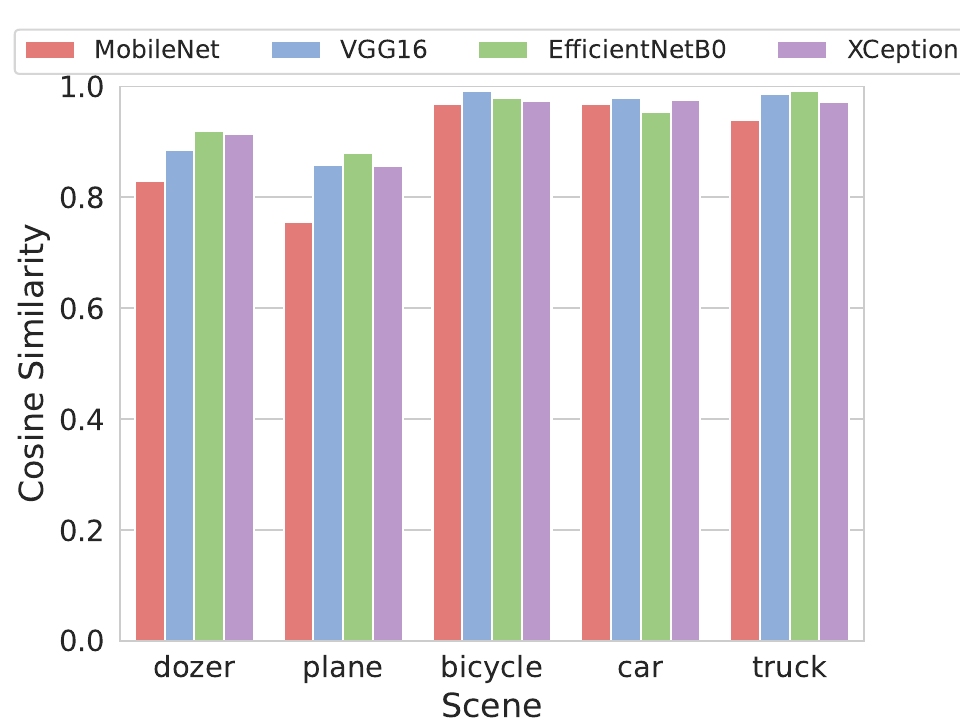}
    \caption{Avg. Cos. Similarity $\uparrow$}
    \label{fig:rq1-cos-sim}
  \end{subfigure}
  \begin{subfigure}{0.19\textwidth}
    \includegraphics[width=\linewidth]{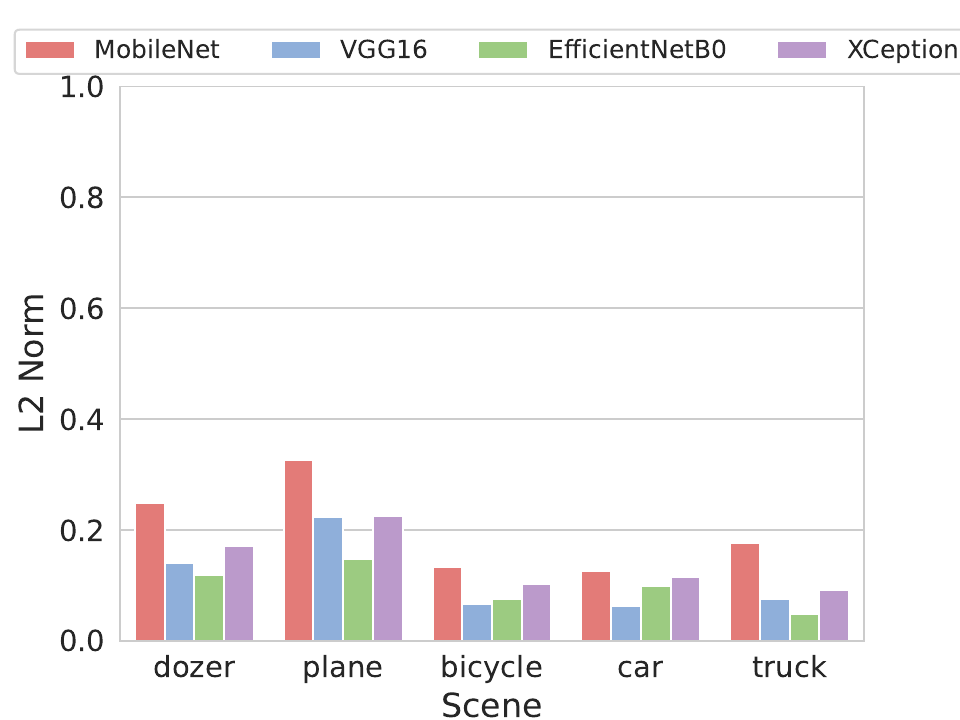}
    \caption{Avg. L2 Norm $\downarrow$}
    \label{fig:rq1-l2norm}
  \end{subfigure}
    \begin{subfigure}{0.19\textwidth}
        \includegraphics[width=\linewidth]{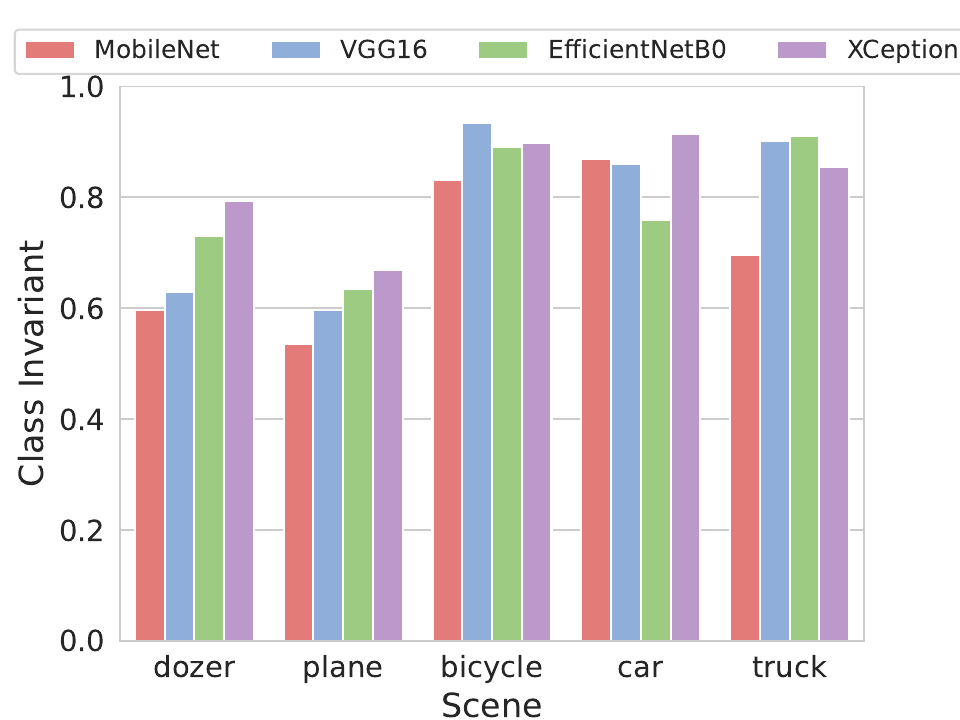}
    \caption{Class Invariance $\uparrow$}
    \label{fig:rq1-class-invariance}
    \end{subfigure}
  \begin{subfigure}{0.19\textwidth}
    \includegraphics[width=\linewidth]{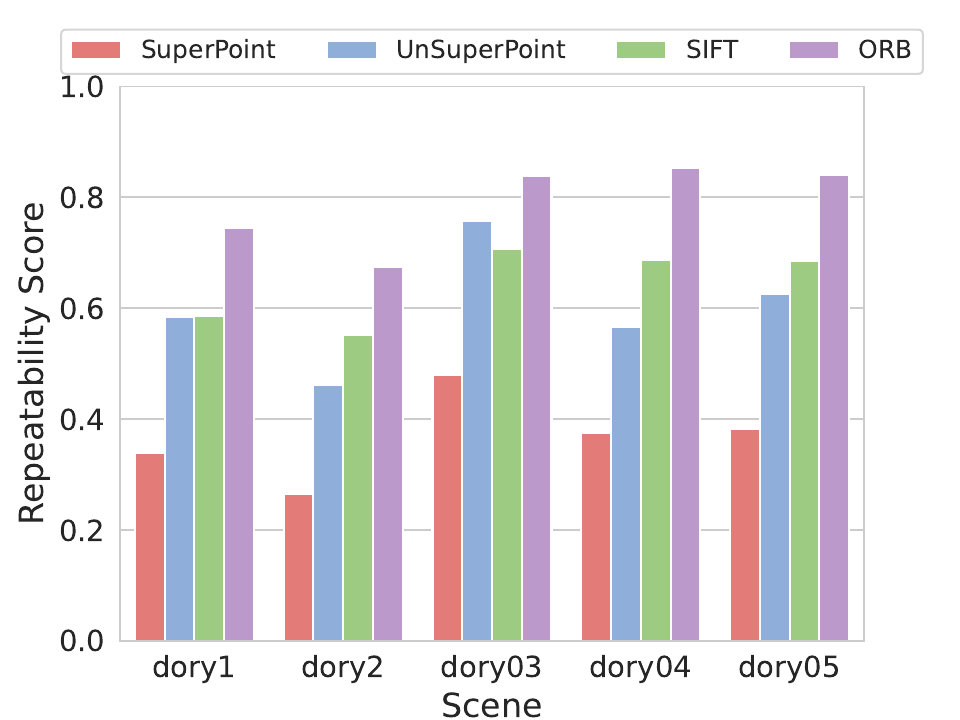}
    \caption{Avg. Repeatability $\uparrow$}
    \label{fig:rq1-repeatability}
  \end{subfigure}
  \begin{subfigure}{0.19\textwidth}
    \includegraphics[width=\linewidth]{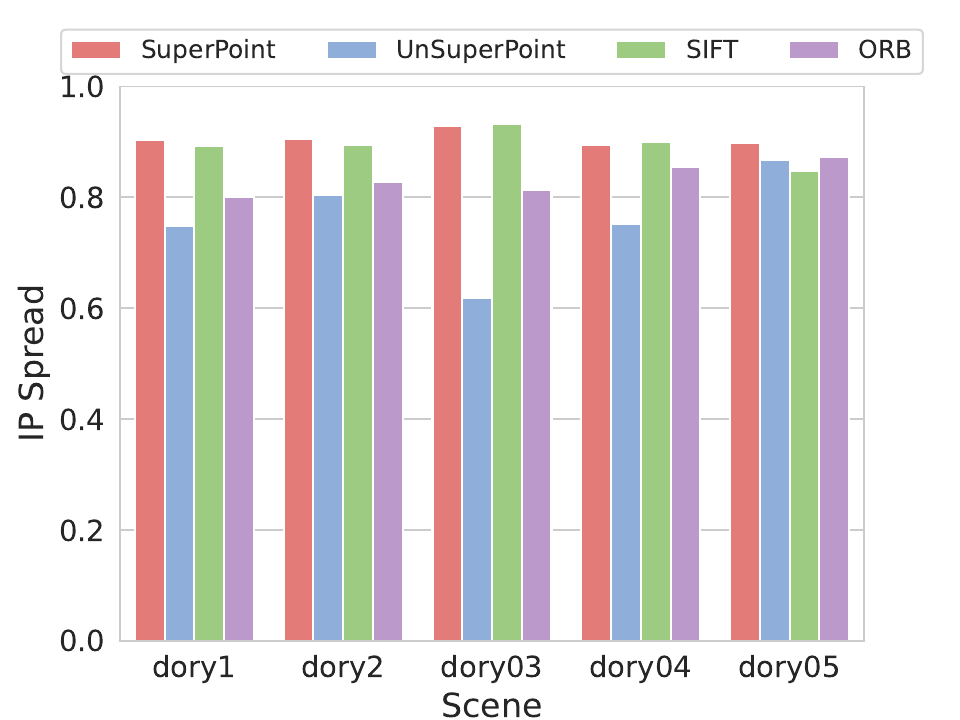}
    \caption{Avg. IP Spread $\uparrow$}
    \label{fig:rq1-ip-spread}
  \end{subfigure}
    \vspace {-.2 \baselineskip}
  \caption{Performance comparison of SUTs under the reality-to-NeRF domain shift in pose-equivalent images in $S_\textrm{real}$ and $S_\textrm{nerf}$.}%
  \label{fig:rq1-all-sut-metrics}
\end{figure*}

\paragraph{SUT Performance under reality-to-NeRF domain shift} We investigate the sensitivity of each SUT to the domain shift from real to NeRF-generated images according to the selected SUT metrics. We consider the set of real images $S_\textrm{real}$ and their pose-equivalent counterparts in $S_\textrm{nerf}$ (e.g. \cref{fig:real-vs-nerf} first column compared to second column). The significance of the SUT metric values under the domain shift is twofold; it can be seen as 1) a measure of quality in the NeRF reconstruction and realness of NeRF-generated images, and 2) an indication of relative performance of each SUT under the reality-to-NeRF domain shift. See \cref{fig:rq1-all-sut-metrics} for average values of each SUT metric in their respective UAV/AUV application. If we passed the exact same image to any SUT, we would expect a value of 1 for cosine similarity, class invariance, repeatability and IP spread, and 0 for the L2 Norm. Any deviation of these values can be seen as an indication of the two properties described above. For each UAV scene, MobileNet shows the lowest scores throughout 13/15 settings compared to other SUTs. ORB is the IPD with the most consistent interest points under the domain shift in all AUV scenes according to the repeatability metric. The interest points of SuperPoint and SIFT appear to find interest points in the most consistent areas compared to the other SUTs. Thus, depending on the vSLAM implementation of choice, we can now make more informed decisions on which IPD or classifier to employ to improve performance downstream tasks with more robustness.
\looseness -1

\paragraph {Statistical Dependence between Image Quality Metrics and SUT Metrics} We are curious whether the image quality metrics PSNR, SSIM, LPIPS have correlation with the UAV/AUV-associated SUT metrics under the reality-to-NeRF domain shift. We consider the real images $S_\textrm{real}$ and their pose-equivalent images in $S_\textrm{nerf}$. Intuitively, the better an image $i_\textrm{nerf}$ is reconstructed, i.e. has better values for PSNR, SSIM and LPIPS, then the features between pairs of $i_\textrm{real}$ and $i_\textrm{nerf}$ should also have higher agreement, and therefore higher SUT metrics. We find there is significant dependence for 38 out of 48 SUT metric / image quality metric combinations, based on Spearman rank correlations up until a p-value of 0.05.%
\looseness -1

\subsection{RQ2: Diversity}

\begin{figure*}[htbp]

  \centering

    \begin{subfigure}{0.24\textwidth}
    \includegraphics[width=\linewidth]{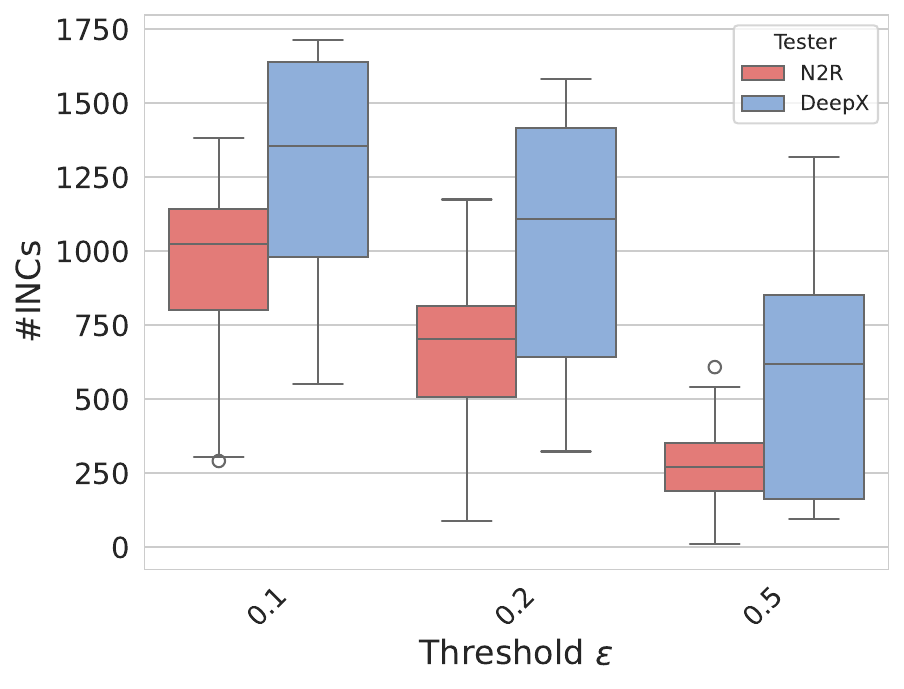}
    \caption{Cos. Similarity}
  \end{subfigure}
  \begin{subfigure}{0.24\textwidth}
    \includegraphics[width=\linewidth]{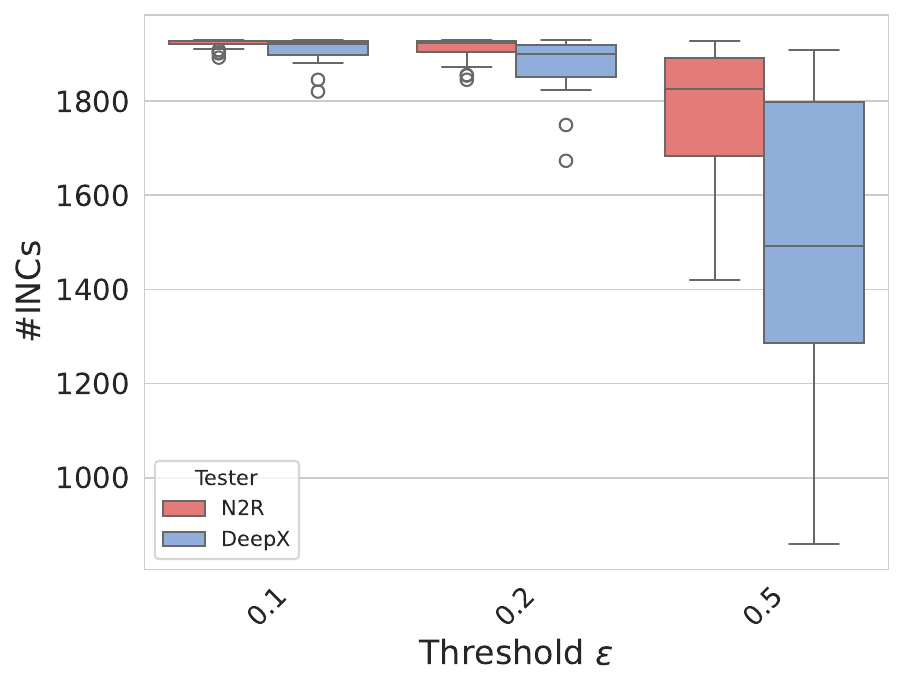}
    \caption{L2 Norm}
  \end{subfigure}
  \begin{subfigure}{0.24\textwidth}
    \includegraphics[width=\linewidth]{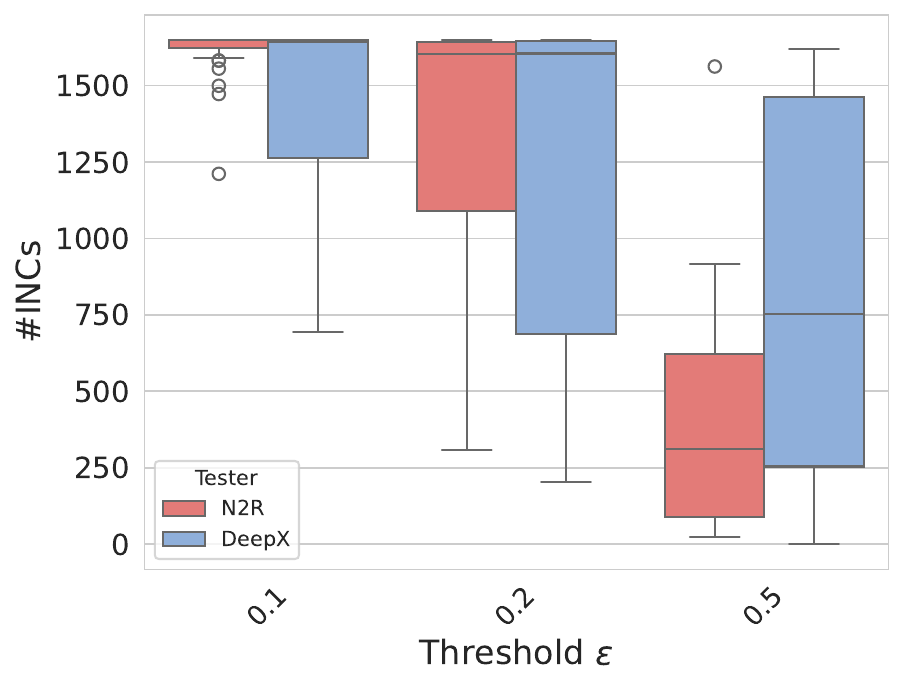}
    \caption{Repeatability}
  \end{subfigure}
  \begin{subfigure}{0.24\textwidth}
    \includegraphics[width=\linewidth]{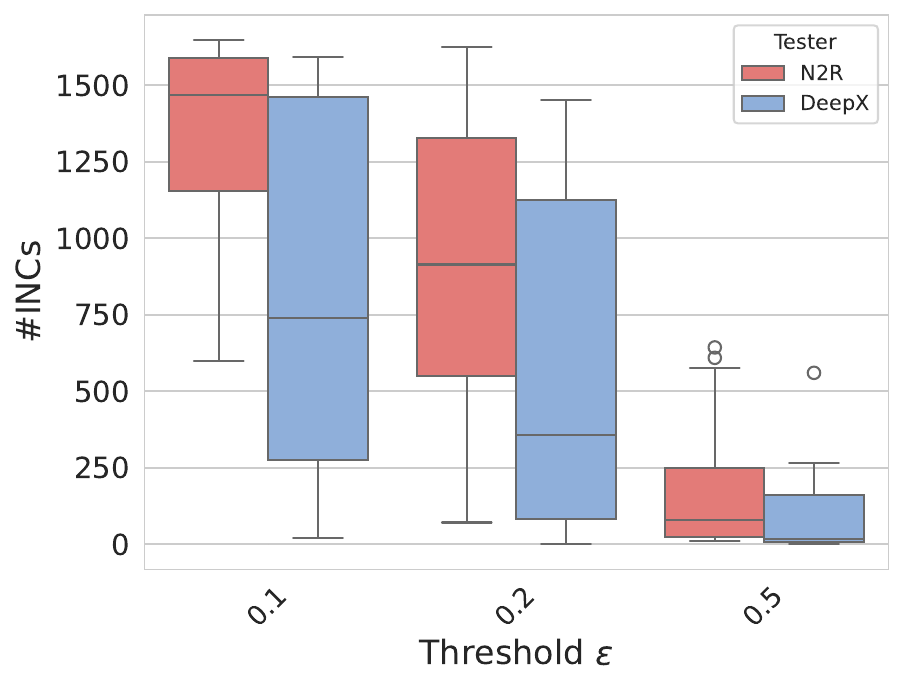}
    \caption{IP Spread}
  \end{subfigure}
    \vspace{ - .5 \baselineskip }
    \caption{Number of inconsistencies per metric under N2R transformations (red) and DeepXplore mutations (blue).}
  \label{fig:rq2-all-incs-per-sut}
   \vspace{ - 0.8 \baselineskip }

\end{figure*}

\begin{figure}
  \begin{subfigure}{0.24\textwidth}
    \includegraphics[width=\linewidth]{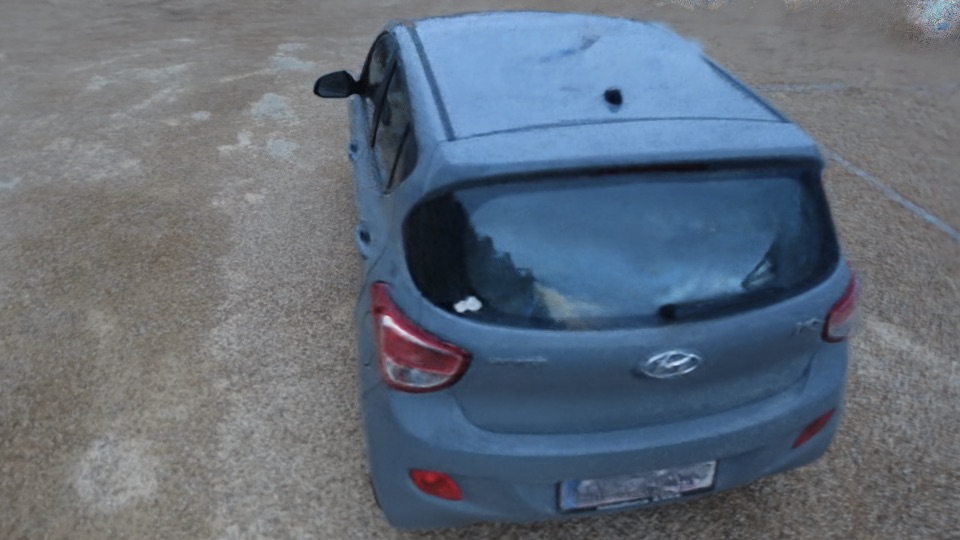}
    \caption{}
    \label{fig:rq2-inc1}
  \end{subfigure}
  \begin{subfigure}{0.24\textwidth}
    \includegraphics[width=\linewidth]{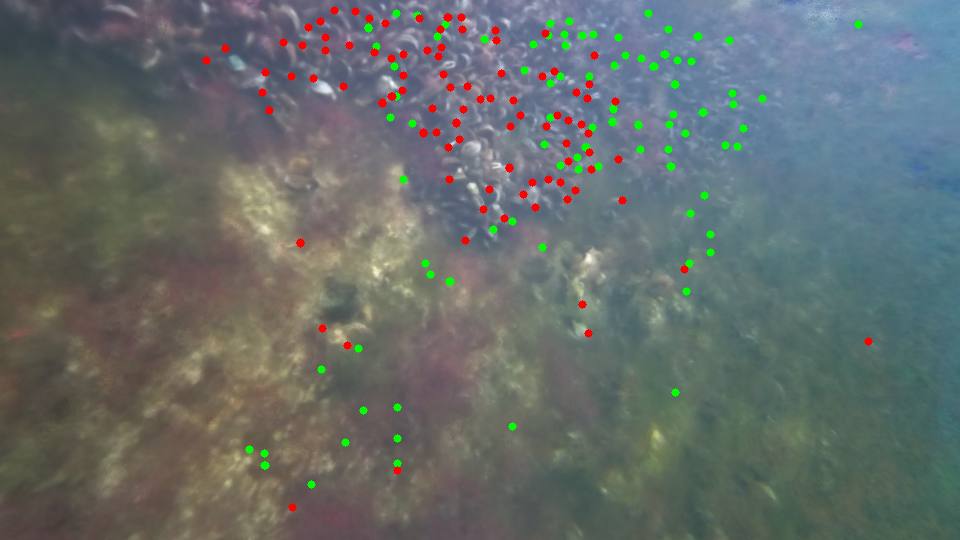}
    \caption{}
    \label{fig:rq2-inc2}
  \end{subfigure}

  \caption{Examples of inconsistent SUT behaviors: Image (a) transformed by $\tau_4$ is classified as ImageNet1k class label ``trash can'' by MobileNet with 0.53 confidence while having average image metric values (PSNR 15.77, SSIM 0.45, LPIPS 0.72). Image (b) transformed by $\tau_4$ has a low repeatability score of 0.37 on SuperPoint. The low score is caused by a low agreement of points from the associated real image (green) compared to points found in this image (red). Image (b) has good values for PSNR 35.12, SSIM 0.70, LPIPS 0.60.}%
  \label{image:sample-images-incs}

\end{figure}

In \cref{fig:rq2-all-incs-per-sut} we show the number of inconsistencies across all SUTs according to our MT framework with three different threshold values for $\epsilon = [0.1, 0.2, 0.5]$. This means that we consider a deviation of a SUT metric by more than 10\%, 20\%, 50\% an inconsistency in the behavior of a SUT. The strictest test is $\epsilon= 0.1$. N2R manages to trigger a higher median of inconsistencies for L2 Norm and IP Spread than DeepXplore mutations, as well as better performance on the strictest tests with $\epsilon = 0.1$ according to the repeatability metric. See \cref{image:sample-images-incs} for two examples of inconsistent SUT behavior.
\looseness -1

\subsection{RQ3: Efficiency}

\noindent
All NeRF trainings and image renderings are performed on two NVIDIA GPUs of type RTX 3090 Ti, averaging $\sim 1.5$ hours of training per model (see \cref{tab:data_info}). Once we have all rendered images of the scenes, the efficiency of executing N2R-Tester is determined by the efficiency of the SUTs processing times. Thus we consider it most appropriate to report the time taken for N2R-Tester to generate test images.
\looseness -1

We report the average FPS for a trained NeRF model to render images at various image resolutions in \cref{tbl:rq3-fps}. DeepXplore mutations can be artificially engineered with OpenCV at around 300 FPS. NeRF models still offer an effective shortcut to realistic 3D scene reconstruction compared to simulating a 3D environment, which requires hiring an experienced software engineer and many working hours.
\looseness -1

\begin{table}
    \centering

    \begin{tabular}{
      >{\small}p{18mm}
      >{\small}c
      >{\small}c
      >{\small}c
    }

        Resolution
        & 480 $\times$ 270
        & 960 $\times$ 540
        & 1920 $\times$ 1080
        \\\hline
        \\[-2mm]

        FPS
        & 4.3 & 2.4  & 0.6
        \\
    \end{tabular}

    \smallskip

    \caption{Rendering frame rates for a trained NeRF model at different image resolutions in frames per second (FPS) $\uparrow$.}%
    \label{tbl:rq3-fps}
\vspace {-1.2 \baselineskip}
\end{table}


\subsection{Threats to Validity and Limitations}

\begin{figure}
    \centering

  \begin{subfigure}{0.45\linewidth}
    \includegraphics[width=\linewidth]{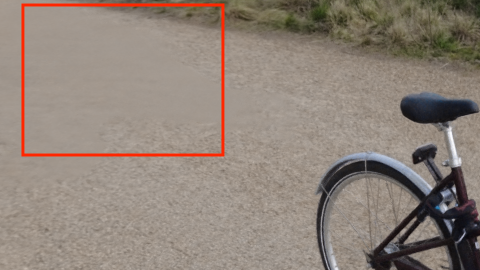}
  \end{subfigure}
  \begin{subfigure}{0.45\linewidth}
    \includegraphics[width=\linewidth]{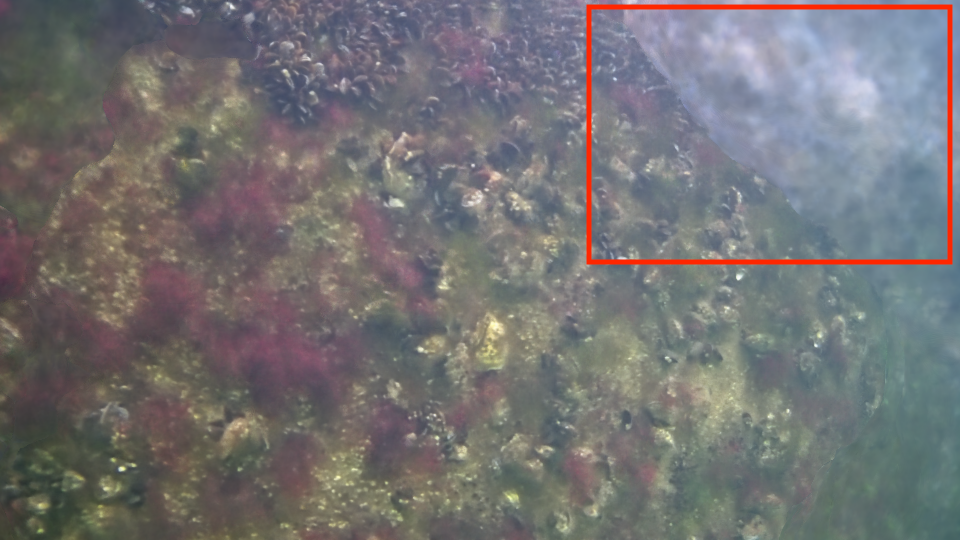}
  \end{subfigure}

  \caption{Left: An example of an under-defined pixel patch in \nfmd{bike}. Right: An example of a ghost wall in \nfmd{dory1}.
  \looseness -1}%
  \label{image:threats}
 \vspace{-1.2 \baselineskip}
\end{figure}

\paragraph {Internal Validity}

For AUV wall inspection, we configured the IPDs to produce hundred highest-confidence interest points (for performance reasons). IPDs can return more points, and it is possible that the behavior of an IPD on NeRF-generated and real images differs further on the omitted points. However, as the points are selected by confidence levels, the difference on the tail of the point distribution should be of decreasing importance.
\looseness -1

\paragraph {External Validity}
The test results with NeRF-generated data for IPDs and classifiers do not necessarily transfer to other scenes, other IPDs, other classification domains, and other classification models. So far the only way to increase confidence of transferability is to accumulate experience with more domains, more models, and more vision components. In our experiments, we have used two vastly different visual tasks (classification and interest-point detection) and used data sets from different domains and different locations. We have used both DNN-based and non-DNN-based IPDs, and large and small classifier models, to increase generalization.
\looseness -1

The NeRF reconstruction process is initialized with random pixels, which gradually converge to a geometrically consistent scene during training. During inference this presents an advantage; NeRFs fail gracefully with ambiguity in the reconstruction, when rendering an image at a view with insufficient radiance samples. This can result in ``artifacts'' or ``ghost walls'', i.e. image patches appearing as random noise or blurry (\cref{image:threats}). Researchers in the NeRF community are in a race trying to outperform each other in mitigating these pixel effects. We are curious as to how exactly these phenomena potentially affect the performance of the SUTs, however this exploration marks a whole line of research in its own right.
\looseness -1

\paragraph {Limitations}
With a trained NeRF model we can only produce test images of one specific instance of a scene or object. Generally, DNNs are trained to classify a wide range of visually distinct instances of the same class. Still, NeRF models can serve as a valuable test image generation technique in scenarios where a particular subset of classes is of interest, for example in the case of transfer learning.
\looseness -1
\section{Related Work}%
\label{sec:relatedwork}

\paragraph{Metamorphic testing for vision components} 
DeepXplore\,\cite{pei2017deepxplore}, DeepTest\,\cite{tian2018deeptest} and DeepRoad \cite{zhang2018deeproad} aim to uncover erroneous behavior of neural networks by synthesizing inputs on a per-frame basis. However they cannot keep geometric properties of a 3D scene or guarantee spatio-temporal consistency. Hence they are unsuitable for testing vSLAM components such as IPDs processing consecutive frames. MT has been applied to evaluate aerial drone control policies in simulation\,\cite{lindvall2017metamorphic}, \cite{adigun2022metamorphic}. Here, the SUT is the drone controller and test inputs are various states of the simulated environment in combination with a mission. Our work addresses testing image processing algorithms with images as test inputs, hence posing as a distinct approach to applying metamorphic testing in the robotics domain. 
\looseness -1

\paragraph {NeRFs in Robotics}
NeRFs can improve vision-based trajectory planners and collision avoidance for aerial drones\,\cite{adamkiewicz2022vision}. NeRF2Real fuses a NeRF with a physics simulator to achieve fast renderings of a scene and modeling of dynamic objects, the robot body, interactions, and collisions\,\cite{byravan2023nerf2real}. SPARTN  is a data augmentation technique for an eye-in-hand robotic arm that feeds NeRF-generated visual samples of grasping behaviors into an imitation learning pipeline\,\cite{zhou2023nerf}. These works improve robotics functionality, however we automate test data generation for vision tasks in robotics. 
\looseness -1

\paragraph {NeRFs for high quality rendering}
Several works aim to improve of NeRFs visual clarity caused by inaccurate or inaccessible camera poses during training, e.g. Bundle-Adjusting Neural Radiance Fields\,\cite{lin2021barf} or Loc-NeRF \cite{maggio2023loc}. These methods improve visual pose estimation and navigation through scenes. We use NeRFs as realistic test mechanisms to uncover faulty behaviors of navigation components. Other types of NeRF architectures reconstruct 3D underwater scenes by reducing back-scatter effects \cite{levy2023seathru} or color distortions \cite{zhang2023beyond}.
\looseness -1
\section{Conclusion}
\label{sec:conclusion}

N2R-Tester is a metamorphic testing and test data synthesis method for perception components of autonomous vehicles. Like other metamorphic methods, it does not use domain-specific or project-specific rules to specify tests. N2R-Tester offers a plethora of advantages over other image generation techniques for navigation applications. First, it grants precise control over image transformations, as opposed to models such as GANs. Second, unlike other computer graphics methods, such as textured meshes combined with sophisticated rendering engines, it does not require explicit 3D models of the scene. A NeRF model is a continuous function approximation capable of implicitly representing complex scenes with arbitrary shapes, appearances and textures, which offers great flexibility over traditional techniques. Third, it does not require human annotation. Finally, N2R-Tester can be potentially extended to test stereo-vision, as it can render two offset camera views. NeRFs inherently model depth in the $\alpha$-channel of the output layer and could therefore also be used to test depth perception.
\looseness -1

\subsubsection*{Acknowledgments}
This project has received funding from the European Union’s Horizon 2020 research and innovation programme under the Marie Skłodowska-Curie grant agreement No 956200.
Fragments of \cref{fig:nerf-overview} are generated with DALL-E by Open AI.
\looseness -1

\bibliographystyle{IEEEtran}
\bibliography{lib/nerf, lib/test, lib/misc}

\end{document}